# Literature-based Discovery for Landscape Planning

**David Marasco[1], Ilya Tyagin[2], Justin Sybrandt[3], James H. Spencer[4] and Ilya Safro[2]**

[1] Planning, Design and the Built Environment, Clemson University, Clemson SC, USA
[2] Department of Computer and Information Sciences, University of Delaware, Newark DE, USA
[3] School of Computing, Clemson University, Clemson SC, USA
[4] School of Architecture, Louisiana State University, Baton Rouge LA, USA

Corresponding author: David Marasco, dmarasc@clemson.edu

## Abstract

This project demonstrates how medical corpus hypothesis generation, a knowledge discovery field of AI, can be used to derive new research angles for landscape and urban planners. The hypothesis generation approach herein consists of a combination of deep learning with topic modeling, a probabilistic approach to natural language analysis that scans aggregated research databases for words that can be grouped together based on their subject matter commonalities; the word groups accordingly form topics that can provide implicit connections between two general research terms. The hypothesis generation system AGATHA was used to identify likely conceptual relationships between emerging infectious diseases (EIDs) and deforestation, with the objective of providing landscape planners guidelines for productive research directions to help them formulate research hypotheses centered on deforestation and EIDs that will contribute to the broader health field that asserts causal roles of landscape-level issues. This research also serves as a partial proof-of-concept for the application of medical database hypothesis generation to medicine-adjacent hypothesis discovery.

## Funding

This research was funded by National Science Foundation grant numbers 1633608 and 2027864. The authors express their gratitude to the NSF for its generous support of their research.

## <u>Introduction</u>

The recent COVID-19 crisis has put the issue of emerging infectious diseases (EIDs) back in the global spotlight. Addressing EIDs going forward will require widespread interdisciplinary cooperation, as discouraging them is a multifaceted and omnipresent endeavor. Biologists and health experts have frequently asserted that landscape-level issues drive EIDs (e.g. Wilcox and Colwell, 2005; Kapan et al., 2006; Neiderud, 2015), but few landscape-level empirical studies have been conducted (e.g. Spencer et al., 2020, under review). Landscape planners certainly have a role to play in this interdisciplinary effort to better understand the origins of and solutions for fighting EIDs, and in order to carve out their contributions, they must understand more precisely and systematically how the existing empirical literature on EIDs has indicated the role of landscape-level factors. One way in which such research angles can be unearthed is through computer-aided hypothesis generation, which is a series of processes to arrange and analyze text data within one or more literature corpora (Spangler, 2016).

Given the general suggestions that landscape factors are a driver of EID evolution, but little direct empirical evidence, hypothesis generation based on the existing medical corpus of research is the most appropriate starting point for determining how landscape planning can shed further light on EIDs. Most existing scholarship on EIDs originates in the related health, medical, and biological fields, and hypothesis generation from a medical corpus would yield some topics that are also relevant to planning disciplines. In sum, this paper uses the hypothesis generation system AGATHA (Sybrandt et al., 2020) to present results that are logically consistent in terms of having a range of interrelated topics that contained terms relevant to both EIDs and deforestation. It further illustrates the process of using the new AGATHA graphical user interface (GUI) to generate models from which promising research hypotheses based on the latent relationships in the existing literature can be formulated.

## **Literature Review**

Infectious diseases are not new; they have afflicted mankind and other animals since time immemorial. However, technology and the advent of modern medicine have allowed humans to ascertain the causes of these diseases and combat their manifestation. Historically, urban and regional planning has been at the forefront of developments that contributed to disease control, as exemplified by the urban mapping that led to the understanding of the cause of the cholera epidemic of 1854 in London (a communal water source) (Johnson 2006).

The 21st century has experienced the persistent presence of EIDs (Morens and Fauci, 2013), and these should be of increasing concern to planners because the world is becoming more and more interconnected and urbanized. Planning scholars have begun to focus on this new category of disease as a threat of global pandemic and the importance of urban development (Spencer et al., 2020, under review; Hamidi et.al., 2020).  In 2018, 55.3% of the Earth's population lived in cities; this figure is predicted to grow by around 5% by 2030 (United Nations, 2018). The impact of urbanization is also happening in concert with - and even influencing - anthropogenic deforestation, which is causing habitat modification and destruction and putting more humans in contact with non-human animal disease carriers (Gottwalt, 2013). Thus, landscape planners may have a significant role to play in fighting disease outbreaks, especially in areas where the landscape itself is being modified to the detriment of many forms of life.

The links between deforestation and EIDs have already generated ample discussion, but that discussion has not translated into a battery of specific countermeasures that planners can leverage to combat zoonoses globally. For example, Cameroon is an African nation that has been identified as a zoonosis and deforestation hotbed (Wolfe et al., 2005). The need for landscape planning and management as a remedial practice in Cameroon has been identified, but the researcher who ascertained this also acknowledged that creating the capacity to realize such an initiative is a major challenge (Brown, 2018). At the very least, landscape planners need to have something to say about land use allocations in places like Cameroon, and if it is determined that landscape-level process result in EIDs, then they should also have some jurisdiction over the environmental transformations associated with deforestation to protect human health; this is a view echoed by the United Nations REDD Programme (Hicks and Scott, 2019).

Part of the challenge facing planning scholars, however, is that the field must translate general calls from other disciplines such as public health, medicine, and virology identifying landscape issues as EID drivers, into likely, evidence-based hypotheses. Without a long-standing emphasis in practice or theory to draw from, emerging techniques of computer-aided hypothesis generation can assist in generating research targets for landscape planners. This approach to research creation mines publications for terminology, with the goal of organizing that terminology in such a way that implicit/hidden connections between target topics are revealed (Spangler et al., 2014). AGATHA, an innovative and new hypothesis generation system used in this study, incorporates topic modeling to sort search results into discoverable hypotheses that can be created based on the written material describing prior quantitative analyses of broad subject fields (Sybrandt et al. 2020). One way to understand AGATHA is that it provides a quantitative basis for informed speculation likely to lead to promising hypotheses of latent, formerly unnamed concepts likely to result in empirical findings. Topic modeling is based on the respective probabilities of terms describing a predefined number of unnamed topics (Liu et al., 2016), and it assigns terms to topic "bins" with probability scores that allow the terms to be ranked. AGATHA is suited to hypothesis generation for landscape planning because its foundation in medicine/health means that it is likely to unearth topics and topic relationships that pertain to planning concerns like public health and environmental resilience.

## **UMLS Database**

AGATHA is a new machine learning software that has been developed to scan vast numbers of studies, searching for latent relationships between concepts that are not explicitly identified in the existing literature. The corpus from which its results are drawn is MEDLINE (Sybrandt et al., 2020), the terminology of which is searchable through the use of codes in the online Unified Medical Language System thesaurus (UMLS; Bodenreider, 2004; National Institutes of Health, 2019; see bibliography for links). The UMLS thesaurus contains a wide range of medicine-related terminology, and each term in the thesaurus has been assigned a different seven-digit code with a preceding "C". Because of the various aspects of different concepts, there are many codes in the UMLS that refer to similar topics (e.g., "Severe Acute Respiratory Syndrome" is coded as C1175175, while the pathogen "SARS coronavirus" is assigned C1175743). This quasi-duplication of terms is important in this project because some terms occur less frequently than others within the same general subject (scoring in AGATHA is discussed below). Thus, if a particular term cannot currently be scored by AGATHA, there is a good chance that there is a different form of the term that can be entered to get a result.

**AGATHA - knowledge representation**

AGATHA is the abbreviation for **A**utomatic **G**raph Mining **and T**ransformer-based **H**ypothesis Generation **A**pproach (Sybrandt et al., 2020). The AGATHA system pipeline is based on constructing a large heterogeneous semantic graph, where the primary unit of information is a sentence. All other types of information which are presented in this graph (n-grams, coded terms, lemmas, entities, and predicates) are sentence-dependent and are obtained from processing sentences using state-of-the-art NLP techniques. For the further details of AGATHA's construction pipeline, please refer to the extensive description in Sybrandt et al. (2020).

**AGATHA - description of scoring component**

One of the main AGATHA components is the predictor model. The AGATHA predictor is a transformer-based deep learning model, which is trained to rank published semantic predicates (valid term-term connections) above noisy negative connections. Given a pair of concepts of interest, the AGATHA predictor outputs the likelihood score of this pair of concepts being connected to each other. The AGATHA predictor model is intended to be used as a recommendation system. Given a set of candidates (pairs of concepts of interest), the predictor will calculate the scores for all of the pairs, and pairs with the highest scores can be investigated further by a domain expert, potentially with assistance from the AGATHA topic query component.

**AGATHA - description of topic query component**

The topic query module was designed to allow a domain expert to explore the semantic space to uncover possible connections between concepts of interest through a combination of topic modeling and network analysis. A similar technique was used in the MOLIERE system (Sybrandt et al., 2017), where it has already proven to be effective in real-world medical studies (Sybrandt et al., 2017; Sybrandt et al. 2018; Aksenova et al., 2019). AGATHA builds on the MOLIERE system by employing a predictor model for connections (which the system "learns" from) instead of relying on heuristics.

The topic query module first extracts a reasonably small subset of sentences relevant to the given pair of concepts from the huge semantic network, which contains hundreds of millions of concepts (sentences, n-grams, entities, etc.) and billions of connections between them. This subset is extracted based on a shortest path between the source and target nodes in the network (a pair of concepts of interest).

Next, this set of extracted sentences is used as an input for a topic modeling scheme. The system employs LDA (Latent Dirichlet Allocation) (Blei et al., 2003; Blei, 2012) topic modeling to generate the set of topics occurring in the input dataset. Importantly, we use first-order neighbors of sentences to generate a per-sentence token dictionary. These first-order neighbors include the following token types: coded terms, entities, lemmas, and n-grams. As observed in earlier experiments, different types of tokens imply different contexts and levels of granularity of resulting topics. Lemmas are the most commonly used and are presented in all extracted sentences, whereas other types of tokens (e.g. n-grams) may be absent. For each of these tokens, the AGATHA model contains the unique coordinate of each token in 512-dimension unified semantic space, which is obtained as a result of embedding the large semantic network mentioned previously.

The LDA model generates topics, which are distributions over tokens from a given corpus vocabulary. Topics can be treated as summaries of sentences sharing similar tokens. For each of these topics, we calculate coordinates in a unified semantic space as a weighted sum of corresponding tokens and their probabilities.

As a result, there is a set of topics, source and target concepts, and their corresponding coordinates in a unified semantic space (presented as nodes), which are then used to construct a KNN (K-nearest neighbors) network. To show a potential way the source and target concepts can be connected, a shortest path is constructed in the network. The shortest path traces one set of edges (connections between the nodes) within in the network.

## AGATHA - description of topic model visualization

To make the process of hypothesis exploration interactive and purposeful, the AGATHA Semantic Visualizer (ASV) module was created; it is a graphical interface for the AGATHA topic query component (Tyagin, I. and Safro, I., 2021). It allows a domain expert to adjust the LDA and KNN network parameters and reconstruct the topic network on the fly. It also introduces an LDA biasing technique, which is used to make specific types of terms more pronounced in the resulting topics.

As was mentioned earlier, each token in the vocabulary has a corresponding coordinate in AGATHA unified semantic space. As a result, topic centroid coordinates also have coordinates of the same dimensionality (512). To be able to arrange topics according to their "real" positions in the semantic space, which potentially improves the interpretability of the information presented in the topic network, the dimensionality was reduced to two in the ASV, and the resulting coordinates are used as a layout for the topic network. The ASV uses a PCA technique to perform dimensionality reduction.

The ASV also allows a domain expert to construct multi-nodal paths from a source node to a target node by selecting one intermediate component at a time; the new path that is then traced by the system is a concatenation of two consecutive shortest paths: from a source node to an intermediate component and from an intermediate component to a target node. This auto-generation of new paths is particularly useful when a topic of interest is discovered by the LDA model but is not presented in the shortest path. This structure of ASV conceptual mapping is well aligned with current recommendations within the planning literature recommending greater attention to the complex upstream and downstream events relating EIDs to a sequence of precursor events (Spencer et al., 2020, under review).

Limitations of the UMLS and the AGATHA Model:

1) External: data set limitations - The UMLS thesaurus is a manually curated dataset, which updates only twice a year and some terms cannot be found at a time they are needed.

2) Internal: user interface - The system needs to be retrained regularly to introduce new terms and connections in the knowledge base. But the system cannot be trained in "online" mode (incrementally), so we have to retrain it from scratch every time, which is time consuming.

3) Approach limitations: In scientific literature, negative results tend to be unpublished, which limits us in terms of absence of negative connections and the necessity of introducing synthetic negative samples manually when we train the system. Moreover, some published results are not reliable and trustworthy, and we do not have a metric to properly address this issue.

**Method**

Finding the terms for analysis in this study was not simply a matter of looking up words in the UMLS, as AGATHA's scoring system does not yet have all of the possible UMLS terms integrated into its framework. Furthermore, the nature of the terms being evaluated had to be considered in a somewhat subjective way. The broader a term is, the higher the score it tends to generate in an AGATHA pairing, but the quality of the score value for a very general term is offset by its lack of specificity in association with another term. This leads to modeling that is too broad to be effective in trying to pin down hypotheses.

The other side of the coin is that very specific terms in AGATHA tend to generate lower scores, and the lower an association score is in AGATHA, the more difficult it is to identify meaningful connections in the topic modeling phase. Thus, it is up to the researcher using AGATHA to determine how specific a term needs to be to get useful results. In this project, the researchers matched individual EIDs with terms that describe different aspects of landscape and urban planning. Because AGATHA is still in development, many terms were not present in the scoring database, but most of the EIDs of interest *were* present in one form or another. The other challenge was finding planning-related terms in the UMLS since it is a medical term thesaurus. In spite of this limitation, we were confident that some planning terms would come up in the UMLS because of the connection between public health and planning, especially given the literature identifying landscape-level issues by health and medical researchers.

It should be noted that after a few preliminary searches, COVID-19 terms were removed from consideration because they are too recent to have been fully integrated into the AGATHA framework. COVID-19 is extremely topical, and it has plenty of research avenues to explore because of the unknowns that still surround it. However, asking AGATHA to generate meaningful COVID-19-related connections is not realistic at this point; using other zoonotic EIDs to evaluate AGATHA results for planning is thus the most logical alternative.

The first operation that AGATHA completes is quantifying the relationship between two chosen terms in the literature by generating an AGATHA Score. After multiple searches in the UMLS thesaurus and the generation of AGATHA correlation scores for different term pairs, the most pertinent pairs with high scores were chosen for topic modeling (Table 1):

| UMLS pair | AGATHA score | UMLS EID term | UMLS planning term |
|-----------|--------------|----------------|---------------------|
| C0001175 C0557812 | 0.903238678 | Acquired Immunodeficiency Syndrome | Community college |
| C0001175 C0079201 | 0.8523877025 | Acquired Immunodeficiency Syndrome | Deforestation |
| C1175175 C0079201 | 0.7488061786 | Severe Acute Respiratory Syndrome | Deforestation |
| C0001175 C0442629 | 0.7205444455 | Acquired Immunodeficiency Syndrome | Urban road |
| C0019682 C0557812 | 0.6865320921 | HIV | Community college |
| C0016627 C0079201 | 0.5812933087 | Influenza in Birds | Deforestation |
| C1175175 C0557812 | 0.5711338043 | Severe Acute Respiratory Syndrome | Community college |
| C0029343 C0079201 | 0.5683526397 | Influenza A Virus, Avian | Deforestation |
| C0019682 C0079201 | 0.5403933525 | HIV | Deforestation |
| C1613950 C0079201 | 0.532837379 | Influenza A Virus, H5N1 Subtype | Deforestation |
| C0019682 C0442629 | 0.4990538418 | HIV | Urban road |
| C0029343 C0557812 | 0.4912571251 | Influenza A Virus, Avian | Community college |
| C0016627 C0557812 | 0.4796778619 | Influenza in Birds | Community college |
| C1175175 C0442629 | 0.4699107111 | Severe Acute Respiratory Syndrome | Urban road |
| C0001175 C0032864 | 0.4640626848 | Acquired Immunodeficiency Syndrome | Power Plants |
| C1615607 C0079201 | 0.4637601674 | Influenza A Virus, H1N1 Subtype | Deforestation |
| C1615607 C0557812 | 0.4636913538 | Influenza A Virus, H1N1 Subtype | Community college |
| C0001175 C0562310 | 0.453883487 | Acquired Immunodeficiency Syndrome | Shopping center |
| C0001175 C0087138 | 0.4414945602 | Acquired Immunodeficiency Syndrome | Urban Planning |
| C0206750 C0079201 | 0.4376486719 | Coronavirus Infections | Deforestation |

*Table 1: The top 20 UMLS code combinations with their AGATHA correlation scores and their exact UMLS term labels*

The terms in the planning column were originally selected because they describe various features in landscapes and urban areas. Currently, the AGATHA scoring system does not contain many of the terms in the UMLS, so the search for viable pairs of terms included a broad range of terms related to planning and cities (including businesses and public buildings, as they represent land uses) to obtain a list of viable results.

As can be seen in Table 1, deforestation and community college are the most prevalent planning-related terms in the top 20 AGATHA results. Of the results listed, half are over the 0.5 threshold, but only four are over 0.7; the pairs with scores over 0.7 are considered promising based on their relative rankings and the recommendations of the AGATHA developers. Deforestation appears twice in those top four results, suggesting that it holds potential as a landscape research angle to address different EIDs. Deforestation was thus chosen as the focus planning term for this project. For the models, deforestation is paired with Acquired Immunodeficiency Syndrome (hereafter referred to as "HIV/AIDS"; score: 0.852) and Severe Acute Respiratory Syndrome (hereafter referred to as "SARS"; score: 0.749).

Once the analysis pairs were determined, they were entered into the ASV. When activated, the ASV presents a two-dimensional plot of nodes and edges that represents the pair of terms being evaluated and all of the topics generated by those terms (see Figure 1 for an example). There is one node for each input word/phrase, and the number of topic nodes is determined by the user. As mentioned previously, the ASV allows for an analyst to control several network parameters through the adjustment of sliders in the GUI, and each time one or more parameters is changed, the model reconstitutes itself.

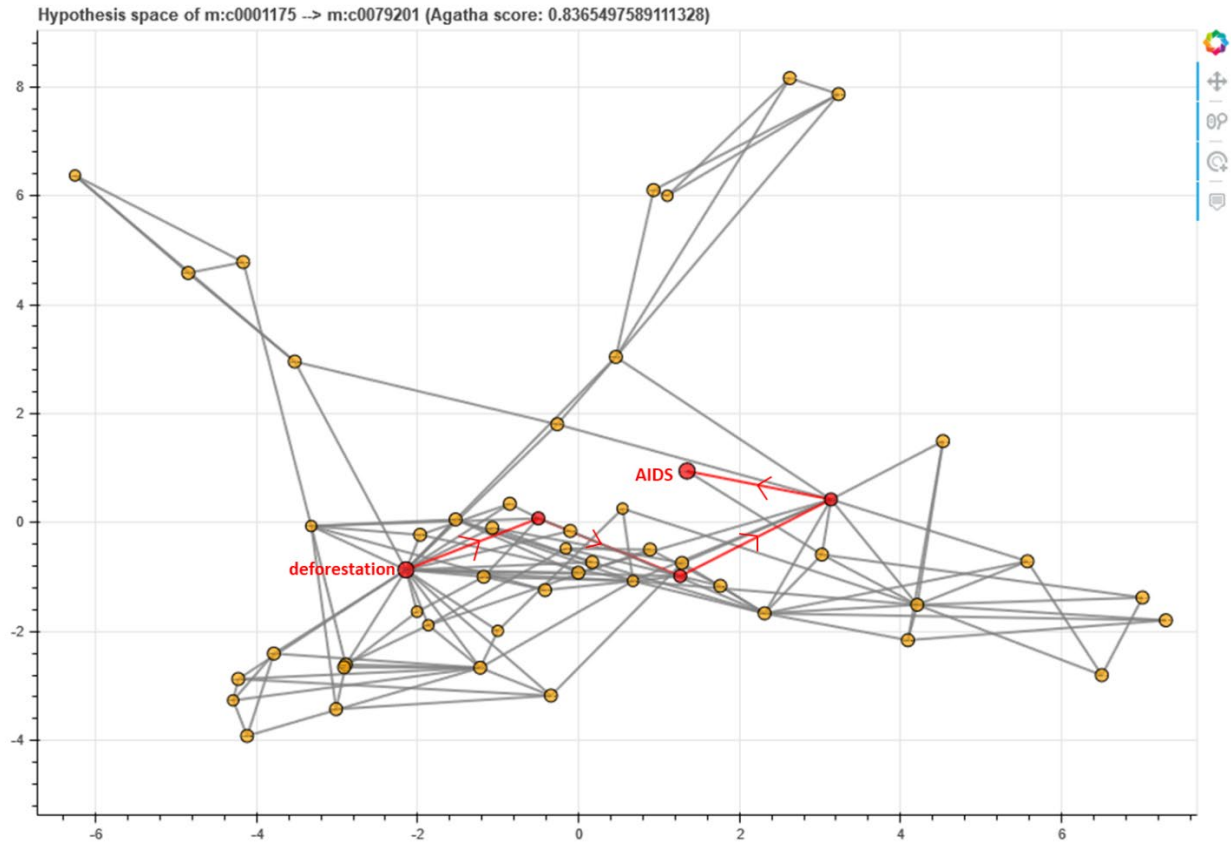

*Figure 1: Two-dimensional representation of the model for AIDS and deforestation with the topic path of interest highlighted in red (50 topics overall). The AGATHA score is close to the one reported in the table, but it is not the same because it is the product of just a single analysis run.*

The ASV quantifies in two-dimensional space the summation of comparative conceptual space between identified potential concepts found in the literature. Thus, the cartesian pathway distance between two nodes is an estimate of the conceptual distance between endpoint and intermediate concepts found in the literature. An ASV presents the shortest conceptual path between the two search terms first, and that path incorporates one extra node as an intermediate concept. This means that the transition between the input terms is dependent on the items in that one node, and repeated trials of the ASV suggest that a single-node transition does not provide enough semantic granularity to hint at one or more hypotheses that conceptually link the endpoints. Since the ASV will draw the shortest path between the endpoints automatically when an inactive node is clicked, it is possible to rapidly find routes that include more than one extra node. This functionality goes beyond that of AGATHA's predecessor MOLIERE, which relied more on term relationship score rankings (Sybrandt et al., 2018). The relationship score between input terms is the starting point for AGATHA hypothesis generation.

The ASV has been designed to be used by people in a variety of disciplines; the term aggregation, sorting, and pathing all takes place "under the hood." However, an interpretation of the ASV results must follow two guidelines: any path that has a portion where an edge was traversed twice is discarded, and any path that appears to deviate significantly from the core cluster of nodes should also be discarded. For example, in Figure 1 it is clear that at least two branches of the node network extend far from the primary cluster, so nodes on those branches were not used for pathfinding because the distance indicates that they are not strongly related to either of the origin nodes.

## Results

### *Model 1: HIV/AIDS and deforestation*

The model for the link between HIV/AIDS and deforestation has 50 topics and a token profile as follows (minimum of 1, maximum of 5): mesh/umls = 4; lemmas = 1; entities = 3; ngrams = 1. MeSH (**Me**dical **S**ubject **H**eadings) (National Institutes of Health, 2020) terms have contextual meaning because they refer to the subject matter of the articles that they describe; they are therefore desirable in a MEDLINE modeling environment. Lemmas are base words that appear in extracted text (Pereira et al., 2008), and they are drawn out with greater frequency; each lemma is labeled with a part of speech that is based on its grammatical context as determined by AGATHA. In this version of the ASV, it is recommended that lemmas be set at "1" for all models, as they occur with high frequency in the topics even when the weight setting is minimized.

To arrive at logical statements, it is assumed that there is a causal direction in the models. Logic suggests deforestation more as a causative agent of EIDs than the inverse, so the analysis proceeds from the deforestation node to the EID node in both of the models illustrated below. Hence, the topics that are in the path in Figure 1 are listed below in Table 2 in the order they proceed from deforestation to HIV/AIDS:

| Topic 35 |
|---|
| l:adj:rural: 0.023 |
| l:noun:community: 0.021 |
| l:adj:clinical: 0.018 |
| m:d004271 - DNA, Fungal: 0.017 |
| e:clinical_trial: 0.016 |
| l:verb:follow: 0.016 |
| l:verb:describe: 0.015 |
| m:d013420 - Sulfamethoxazole: 0.015 |
| m:d014295 - Trimethoprim: 0.015 |
| e:associate_with: 0.014 |
| |
| **Topic 19** |
| m:d015662 - Trimethprim, Sulfamethoxazole Drug Combination: 0.062 |
| m:d000890 - Anti-Infective Agents: 0.055 |
| m:d005740 - Gases: 0.022 |
| m:d008697 - Methane: 0.020 |
| e:no_significant: 0.020 |
| m:d007166 - Immunosuppressive Agents: 0.018 |
| l:noun:lymphoma: 0.015 |
| m:d008081 - Liposomes: 0.015 |
| e:statistically_significant: 0.013 |
| l:noun:aids: 0.013 |
| |
| **Topic 43** |
| m:d000998 - Antiviral Agents: 0.030 |
| e:central_nervous_system: 0.019 |
| m:d017245 - Foscarnet: 0.017 |
| m:d007070 - Immunoglobulin A: 0.017 |
| l:noun:malaria: 0.017 |
| l:noun:hiv: 0.016 |
| m:d016053 - RNA, Protozoan: 0.014 |
| l:adj:central: 0.013 |
| l:noun:aids: 0.012 |
| m:d019476 - Insect Proteins: 0.012 |

*Table 2: The topics in the preferred AIDS-deforestation model. Each topic has the top ten most probable terms listed. Key: l - lemma, m - mesh/uml, e - entity, n - ngram*

The top ten terms are listed with their respective token types, parts of speech (only for lemmas), and probability scores (all term scores in the topic sum to one); the MeSH terms have a catalog number attached to them as well. As expected, topic 35 (the topic closest to the deforestation node) has some terms that pertain to planning, while topic 43 (the topic closest to the HIV/AIDS node) is composed almost solely of medicine-related terms. On the information in the undefined topics, it proposed that the chain flows from deforestation to AIDS as follows:

*fungal shifts in rural communities → methane releases from deforestation → disease treatment*

Note that the events in the chain are not necessarily causative relative to each other, and in particular it should not be assumed that non-proximal nodes have logical/immediately identifiable conceptual links. Methane gas releases do not necessarily cause infectious diseases directly, for example; however, the presence of such releases might be an indicator of end-stage deforestation that leads to species displacement and zoonotic disease transmission.

Current research corroborates the connections in the chain. In investigating the use of tropical land as palm and rubber plantations, Brinkmann et al. (2019) found that a primary impact of the conversion of rainforest land to plantations is the change in the fungal content of soils, which in turn catalyzes ecosystem alteration. More recently, Kroeger et al. (2020) discovered that livestock pastures in the Amazon (which are essentially completely deforested) contain elevated levels of methanogens, thus accounting for the abundance of methane releases from these land uses. Notably, the study by Kroeger does not assign a specific cause to the proliferation of methanogens in pasture soil, so research remains to be done on the factors that drive it.

### *Model 2: SARS and deforestation*

The second model shows a potential relationship between deforestation and SARS, and suggests the same direction-of-effect logic as the deforestation-HIV/AIDS model (Figure 2). It has two topics that form the connection between the end nodes, and it maintains the approach of keeping the path nodes within a focused space around the input terms.

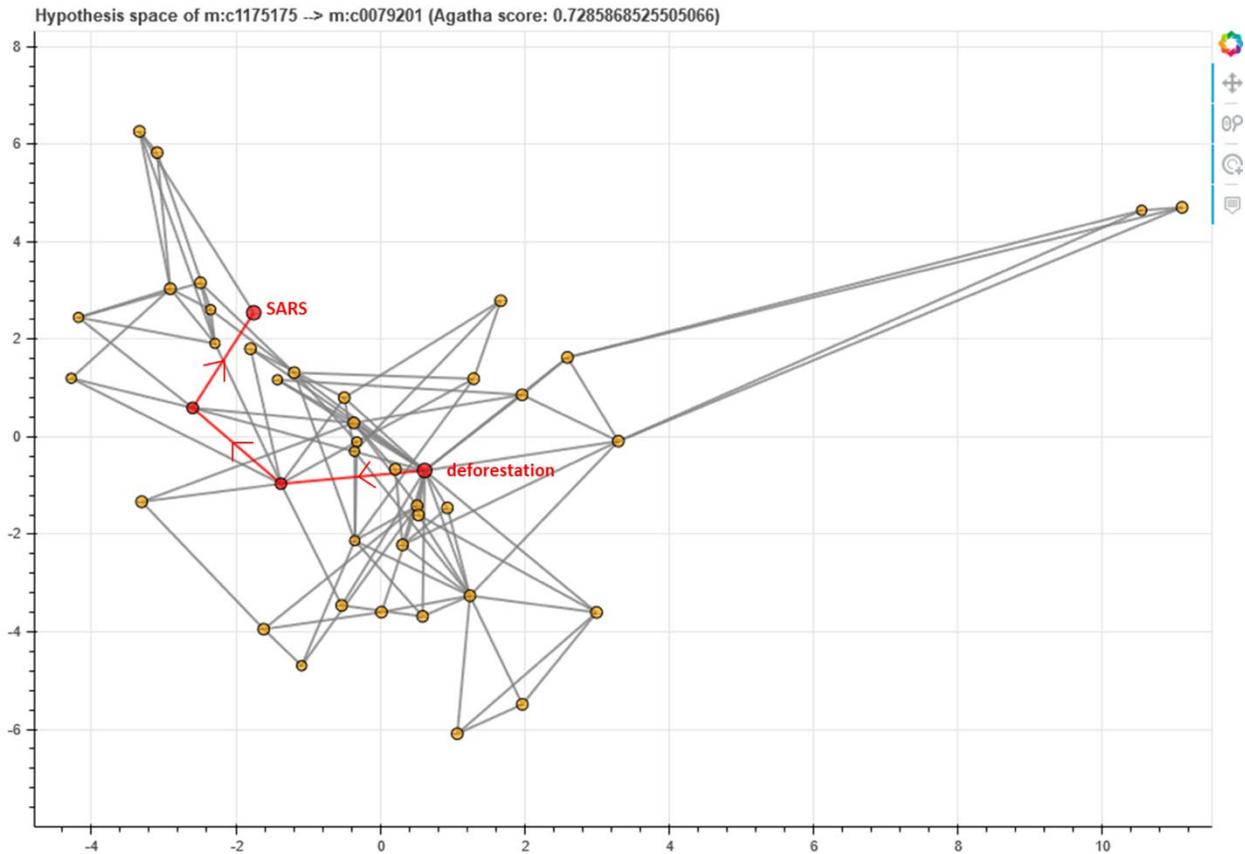

*Figure 2: Two-dimensional representation of the model for SARS and deforestation with the topic path of interest highlighted in red (40 topics overall). Again here, the AGATHA score is slightly different because it comes from a single modeling run in the visualizer.*

This model shares the same token weighting scheme with the deforestation-HIV/AIDS model, but it employs 40 topics instead of 50. Finding planning-related topics that existed on viable paths proved challenging, and this might have something to do with the fact that the AGATHA score for this model is around 0.1 lower than the score for the deforestation-HIV/AIDS one. In the model, topic 27 is the one proximate to deforestation (Table 3).

| Topic 27 |
|---|
| m:d015850 - Interleukin-6: 0.037 |
| m:d010573 - Pesticide Residues: 0.025 |
| m:d016209 - Interleukin-8: 0.023 |
| e:gene_expression: 0.021 |
| m:d012324 - RNA Replicase: 0.017 |
| m:d007306 - Insecticides: 0.013 |
| l:noun:expression: 0.013 |
| e:amazon_basin: 0.012 |
| l:noun:malaria: 0.012 |
| m:d010126 - Ozone: 0.011 |
| |
| **Topic 17** |
| m:d016756 - Immunoglobulins, Intravenous: 0.029 |
| m:d014157 - Transcription Factors: 0.023 |
| m:d001426 - Bacterial Proteins: 0.019 |
| m:d015534 - Trans-Activators: 0.018 |
| m:d016898 - Interferon-alpha: 0.015 |
| m:c088021: 0.014 |
| m:c095698: 0.012 |
| m:d014527 - Uric Acid: 0.010 |
| m:d003278 - Contraceptives, Oral, Hormonal: 0.010 |
| m:d008775 - Methylprednisolone: 0.010 |

*Table 3: The topics in the path from deforestation to SARS. Each topic has the top ten most probable terms listed. Key: l - lemma, m - mesh/uml, e - entity, n - ngram*

Topic 27 suggests that pesticides and insecticides are related to deforestation as a byproduct, and their usage might be found in the Amazonian basin. This yields a concise possible path to an infectious disease:

*pesticide usage on deforested lands → disease manifestation/medicinal treatment*

According to Catacora-Vargas et al. (2012), the expansion of soybean farming in the "Southern Cone" of South America has exacerbated deforestation, and the soybean farming uptick has coincided with a significant increase in the use of pesticides in that area, glyphosate in particular. Glyphosate is a chemical that is known to cause serious biological and environmental disruptions (Toretta et al., 2018), and it is therefore an ecosystem hazard. Ecosystem disruptions lead to the relocations of animal populations, and so pesticide usage in the wake of deforestation can be seen as a potential catalyst for eventual zoonotic disease transmission.

## Discussion: Latent relationships between EIDs and landscape-level planning

This topical research on planning and EIDs represents the first use of the newly constituted ASV module in research. Consequently, finding hypotheses, particularly those that pertain to fields other than medicine - especially one like planning that emphasizes different spatial scales than medicine - requires some trial and error. However, the occurrence of topics that provide promising avenues for planning scholarship to engage the pressing and numerous unknowns related to EIDs more than justifies the application of AGATHA to landscape planning, no matter how nascent the form.

Model 1 produced the more compelling result in this research, due to the multi-stage nature of its causal chain. Methanogens don't cause EIDs directly, but they are an indicator of pasture land use (Kroeger et al., 2020), and many pastures exist as a result of deforestation. The soil conditions that precede methanogen increases are not yet known, but model 1 suggests that an examination of fungal profiles of pre-methanogenic and methanogenic soils may reveal fungal composition shifts as an intermediate landscape phenomenon related to deforestation. Scholarship resulting from this line of inquiry, for example might eventually reconsider the land uses that contribute to fungal profile shifts and ask whether or not those particular land uses are associated with the eventual emergence of zoonotic infectious diseases. Beyond research, finding a relationship between deleterious fungal changes in soil and disease outbreaks would give planning practice and policy  the opportunity to remedy conditions that might lead to zoonotic transmission before they can reach a critical stage - which in many cases of EIDs are human pandemic stages well beyond effective human responses, as has been the case with COVID-19, Ebola, and others.

Model 2 yields a more straightforward proposition with regards to planning: that large-scale agricultural pesticide usage might be associated with the creation of zoonoses. In at least South America, deforestation has often signaled an increase in pesticide use (Catacora-Vargas et al., 2012), so environmentalists and planners alike should determine whether or not pesticide use is actually a crucial link in the zoonosis upstream-downstream sequence of events as described in Spencer et al. (2020, under review). If it is, landscape planners need to find ways to discourage it through prohibitions on certain land use types and the promotion of sustainable and environmentally sound farming practices.

***Example of a landscape-level hypothesis generated from AGATHA:***

*There is a significant relationship between deforestation-driven fungal changes in soil and the manifestation of zoonotic infectious diseases.*

Investigation of this hypothesis would begin with identifying the fungal composition changes that can result from deforestation. Brinkmann et al. (2019) found that fungal shifts occur in tropical areas, so information on these kinds of changes post-deforestation in other climates would be necessary. Such information would be obtainable through cooperation with local agricultural agencies in areas that have recently experienced large-scale deforestation; identifying these areas would require a GIS analysis of various locations around the globe.

The outcome variable in the relationship would be the occurrence of a zoonotic disease proximate to a deforestation site (which would require its own definition). Data on zoonoses is obtainable through global and national health agencies, and the respective times of disease detection would have to be compared with deforestation periods to make any sort of causal inference (i.e., a zoonotic disease that occurred prior to a deforestation event would not be considered a "yes" result for a binary dependent variable). Modeling would proceed with fungal changes defined in one or more independent variables, depending on the variety of fungi and fungal transitions present in different soils. The confirmation of a relationship between fungal shifts and zoonotic disease occurrence would necessitate further research into intervening variables and upstream analysis of how fungal profiles can be controlled and/or discouraged through landscape management. The net result of the effort would be the interaction of planners, agricultural scientists, and health officials to gain more granular insight into the specific mechanisms present in deforestation-driven zoonoses.

## Conclusion: The use of AGATHA methods for landscape-level planning

From the outset, there were inherent limitations in this research that should be borne in mind in future applications of AGATHA. For one, AGATHA uses a medical research corpus, whose original purpose was to derive hypotheses for medicine and health care. Nevertheless, considering the results generated by AGATHA in this research, it is clear that the methods applied here have potential for useful medicine-adjacent hypothesis generation as well; as an emergent method for hypothesis generation, the flexibility demonstrated by this application to a medicine-adjacent field will only improve as AGATHA gets refined and users in different disciplines become familiar with its parameters. In particular, the expansion of AGATHA's terminology set stands to be a transformative advancement for AGATHA that would open up more directly relevant hypothesis generation of landscape-level drivers of EIDs. Specifically, such an advancement would provide preliminary relationship scores (a critical component) for the vast majority of the terms in the UMLS, and thereby expand the research tools for understanding the very mysterious origins of EIDs such as COVID-19.

Hypotheses implied by the connections in AGATHA in this project merit investigation for two reasons: 1) the connections they propose reference established research, and 2) they can speak to the efficacy of AGATHA for hypothesis development outside of medicine. That said, failure to confirm the relationships through research would not invalidate AGATHA's applicability across disciplines, as hypothesis inaccuracy could occur for a couple of reasons that have nothing to do with AGATHA's fundamental design: 1) AGATHA is still being expanded as of the publication of this paper, and 2) there is a human component to using AGATHA and interpreting its results.

AGATHA has been in development since the fall of 2019, and it already has many useful features to help domain experts utilize hypothesis generation in their research so that they may ascertain the nature of previously unknown connections. AGATHA developers are currently focusing on increasing the scope of the system in all of its facets (scoring, terminology set, flexibility in the exporting of data, and tailoring of term priorities), and going forward this should provide users more freedom and improve the overall quality of output topics in the ASV module.

There are some clear future research opportunities regarding AGATHA and landscape planning. For one, the hypothetical connections between certain types of events and EIDs can be investigated by landscape planners and environmental scientists, and AGATHA can then be employed in other planning studies to see if it shows hypothesis improvement over time. Planners and virologists can also look into whether fungal profiles of soil and pesticides are linked to specific types of EIDs, something that is implied by fungal profiles being connected to AIDS (an EID transmitted through bodily fluids; HIV.gov, 2019) and pesticides being connected to SARS (a respiratory EID transmitted through exhaled air and fecal matter; WHO, 2020). Through machine learning, AGATHA will also have its search algorithms refined to target certain term groups to increase its efficiency for different user bases, so landscape planners would do well to find out if AGATHA can be modified to meet their specific research needs more rapidly. Ultimately, AGATHA aims to be a valuable tool for both medical professionals and researchers in medical- and public health-adjacent fields like landscape and urban planning.

　　　The greatest finding of this is the demonstrated potential of AGATHA and related methods that draw on databases of existing research to find innovative hypotheses to empirically investigate related to a class of dire human health challenges where little is known about the origins, effective solutions, and implications that occur in very fast increments and at global scales. Such an approach is in line with the thinking of author Anthony Townsend (2013), who advocates for interdisciplinary aptitude and thorough data analysis in forward-thinking planning. To develop such an approach in landscape planning specifically, one of the most important directions for future research is to get all the environmental journals, planning journals, and animal health journals incorporated into the UMLS. Another interesting directions is to consider the analysis of full text papers instead of the abstracts (Sybrandt et al., 2018) and dynamic topic modeling that takes into account evolving topics (Gropp, C. et al., 2019).

**<u>List of Tables</u>**

Table 1: The top 20 UMLS code combinations with their AGATHA correlation scores and their exact UMLS term labels

Table 2: The topics in the preferred AIDS-deforestation model. Each topic has the top ten most probable terms listed. Key: l - lemma, m - mesh/uml, e - entity, n - ngram

Table 3: The topics in the path from deforestation to SARS. Each topic has the top ten most probable terms listed. Key: l - lemma, m - mesh/uml, e - entity, n - ngram

**Tables**

**Table 1**

| UMLS pair | AGATHA score | UMLS EID term | UMLS planning term |
|-----------|--------------|---------------|--------------------|
| C0001175 C0557812 | 0.903238678 | Acquired Immunodeficiency Syndrome | Community college |
| C0001175 C0079201 | 0.8523877025 | Acquired Immunodeficiency Syndrome | Deforestation |
| C1175175 C0079201 | 0.7488061786 | Severe Acute Respiratory Syndrome | Deforestation |
| C0001175 C0442629 | 0.7205444455 | Acquired Immunodeficiency Syndrome | Urban road |
| C0019682 C0557812 | 0.6865320921 | HIV | Community college |
| C0016627 C0079201 | 0.5812933087 | Influenza in Birds | Deforestation |
| C1175175 C0557812 | 0.5711338043 | Severe Acute Respiratory Syndrome | Community college |
| C0029343 C0079201 | 0.5683526397 | Influenza A Virus, Avian | Deforestation |
| C0019682 C0079201 | 0.5403933525 | HIV | Deforestation |
| C1613950 C0079201 | 0.532837379 | Influenza A Virus, H5N1 Subtype | Deforestation |
| C0019682 C0442629 | 0.4990538418 | HIV | Urban road |
| C0029343 C0557812 | 0.4912571251 | Influenza A Virus, Avian | Community college |
| C0016627 C0557812 | 0.4796778619 | Influenza in Birds | Community college |
| C1175175 C0442629 | 0.4699107111 | Severe Acute Respiratory Syndrome | Urban road |
| C0001175 C0032864 | 0.4640626848 | Acquired Immunodeficiency Syndrome | Power Plants |
| C1615607 C0079201 | 0.4637601674 | Influenza A Virus, H1N1 Subtype | Deforestation |
| C1615607 C0557812 | 0.4636913538 | Influenza A Virus, H1N1 Subtype | Community college |
| C0001175 C0562310 | 0.453883487 | Acquired Immunodeficiency Syndrome | Shopping center |
| C0001175 C0087138 | 0.4414945602 | Acquired Immunodeficiency Syndrome | Urban Planning |
| C0206750 C0079201 | 0.4376486719 | Coronavirus Infections | Deforestation |

*Table 1: The top 20 UMLS code combinations with their AGATHA correlation scores and their exact UMLS term labels*

**Table 2**

| Topic 35 |
| --- |
| l:adj:rural: 0.023 |
| l:noun:community: 0.021 |
| l:adj:clinical: 0.018 |
| m:d004271 - DNA, Fungal: 0.017 |
| e:clinical_trial: 0.016 |
| l:verb:follow: 0.016 |
| l:verb:describe: 0.015 |
| m:d013420 - Sulfamethoxazole: 0.015 |
| m:d014295 - Trimethoprim: 0.015 |
| e:associate_with: 0.014 |
| |
| **Topic 19** |
| m:d015662 - Trimethoprim, Sulfamethoxazole Drug Combination: 0.062 |
| m:d000890 - Anti-Infective Agents: 0.055 |
| m:d005740 - Gases: 0.022 |
| m:d008697 - Methane: 0.020 |
| e:no_significant: 0.020 |
| m:d007166 - Immunosuppressive Agents: 0.018 |
| l:noun:lymphoma: 0.015 |
| m:d008081 - Liposomes: 0.015 |
| e:statistically_significant: 0.013 |
| l:noun:aids: 0.013 |
| |
| **Topic 43** |
| m:d000998 - Antiviral Agents: 0.030 |
| e:central_nervous_system: 0.019 |
| m:d017245 - Foscarnet: 0.017 |
| m:d007070 - Immunoglobulin A: 0.017 |
| l:noun:malaria: 0.017 |
| l:noun:hiv: 0.016 |
| m:d016053 - RNA, Protozoan: 0.014 |
| l:adj:central: 0.013 |
| l:noun:aids: 0.012 |
| m:d019476 - Insect Proteins: 0.012 |

*Table 2: The topics in the preferred AIDS-deforestation model. Each topic has the top ten most probable terms listed. Key: l - lemma, m - mesh/uml, e - entity, n - ngram*

**Table 3**

| Topic 27 |
| --- |
| m:d015850 - Interleukin-6: 0.037 |
| m:d010573 - Pesticide Residues: 0.025 |
| m:d016209 - Interleukin-8: 0.023 |
| e:gene_expression: 0.021 |
| m:d012324 - RNA Replicase: 0.017 |
| m:d007306 - Insecticides: 0.013 |
| l:noun:expression: 0.013 |
| e:amazon_basin: 0.012 |
| l:noun:malaria: 0.012 |
| m:d010126 - Ozone: 0.011 |
|  |
| **Topic 17** |
| m:d016756 - Immunoglobulins, Intravenous: 0.029 |
| m:d014157 - Transcription Factors: 0.023 |
| m:d001426 - Bacterial Proteins: 0.019 |
| m:d015534 - Trans-Activators: 0.018 |
| m:d016898 - Interferon-alpha: 0.015 |
| m:c088021: 0.014 |
| m:c095698: 0.012 |
| m:d014527 - Uric Acid: 0.010 |
| m:d003278 - Contraceptives, Oral, Hormonal: 0.010 |
| m:d008775 - Methylprednisolone: 0.010 |

*Table 3: The topics in the path from deforestation to SARS. Each topic has the top ten most probable terms listed. Key: l - lemma, m - mesh/uml, e - entity, n - ngram*

**<u>List of Figures</u>**

Figure 1: Two-dimensional representation of the model for AIDS and deforestation with the topic path of interest highlighted in red (50 topics overall). The AGATHA score is close to the one reported in the table, but it is not the same because it is the product of just a single analysis run.

Figure 2: Two-dimensional representation of the model for SARS and deforestation with the topic path of interest highlighted in red (40 topics overall). Again here, the AGATHA score is slightly different because it comes from a single modeling run in the visualizer.